%% file: egpaper_final.tex
\documentclass[10pt,twocolumn,letterpaper]{article}

\usepackage{iccv}
\usepackage{times}
\usepackage{epsfig}
\usepackage{graphicx}
\usepackage{amsmath}
\usepackage{amssymb}

\usepackage{multirow}

\usepackage{graphicx}
\usepackage{amsmath}
\usepackage{amssymb}
\usepackage{booktabs}

\usepackage{amsmath}

\usepackage{enumitem}

\usepackage[breaklinks=true,bookmarks=false]{hyperref}

\iccvfinalcopy 

\def\iccvPaperID{9554} 

\ificcvfinal\fi

\begin{document}

\title{Local Attention Transformers for High-Detail Optical Flow Upsampling}

\author{
Alexander Gielisse, Nergis Tomen, Jan van Gemert  \\
Computer Vision Lab, Delft University of Technology \\
{\tt\small a.s.gielisse@tudelft.nl, n.tomen@tudelft.nl, j.c.vangemert@tudelft.nl}
}

\maketitle
\ificcvfinal\thispagestyle{empty}\fi

\input{chapters/abstract}
\input{chapters/introduction}

\input{chapters/related_work}

\input{chapters/method}

\input{chapters/experiments}

\input{chapters/discussion}

{\small
\bibliographystyle{ieee_fullname}

\input{egpaper_final.bbl}
}

\end{document}

%% file: chapters/abstract.tex
\newcommand{\citesota}{
\cite{flowformer, gma, flowformer_plusplus, craft, raft, gmflow, separableflow, gmflownet}
}
\newcommand{\citesotanogmflow}{
\cite{flowformer, gma, flowformer_plusplus, craft, raft, separableflow, gmflownet}
}
\begin{abstract}

Most recent works on optical flow use convex upsampling as the last step to obtain high-resolution flow. In this work, we show and discuss several issues and limitations of this currently widely adopted convex upsampling approach. We propose a series of changes, 
in an attempt to resolve current issues. First, we propose to decouple the weights for the final convex upsampler, making it easier to find the correct convex combination. For the same reason, we also provide extra contextual features to the convex upsampler. Then, we increase the convex mask size by using an attention-based alternative convex upsampler; Transformers for Convex Upsampling. This upsampler is based on the observation that convex upsampling can be reformulated as attention, and we propose to use local attention masks as a drop-in replacement for convex masks to increase the mask size. We provide empirical evidence that a larger mask size increases the likelihood of the existence of the convex combination. Lastly, we propose an alternative training scheme to remove bilinear interpolation artifacts from the model output.
Our proposed ideas could theoretically be applied to almost every current state-of-the-art optical flow architecture. On the FlyingChairs + FlyingThings3D training setting we reduce the Sintel Clean training end-point-error of RAFT from $1.42$ to $1.26$, GMA from $1.31$ to $1.18$, and that of FlowFormer from $0.94$ to $0.90$, by solely adapting the convex upsampler.

\end{abstract}

%% file: chapters/introduction.tex
\begin{figure}[t]
    \centering
        \includegraphics[width=\linewidth]{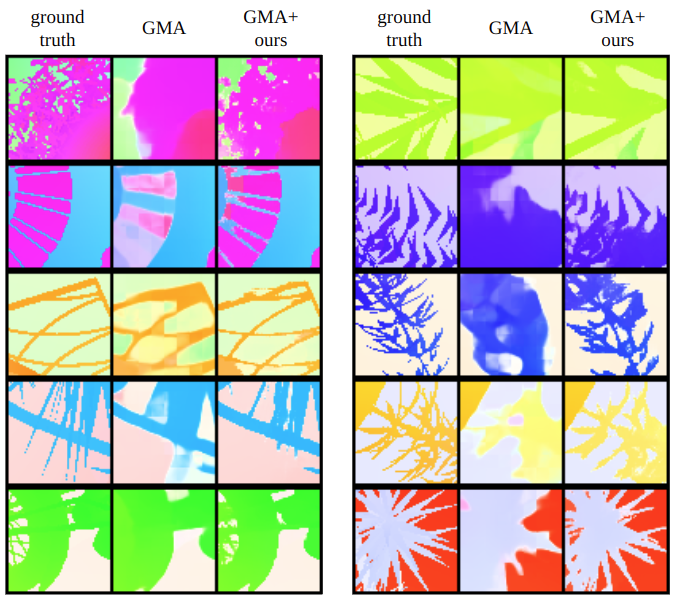}
    \caption{
    Current optical flow methods ignore fine details.  We show 10 High-detail 64x64 patches from the FlyingThings3D \cite{flyingthings} dataset and compare a recent baseline (GMA~\cite{gma}) with adding our upsampler. Our upsampling better preserves fine details. 
    }
    \label{fig:example_patches}
\end{figure}

\section{Introduction}

Current state-of-the-art deep optical flow models \citesotanogmflow{} is heavily inspired by RAFT~\cite{raft}. RAFT applies an iterative, recurrent, optimization and makes this possible by reducing memory and compute by predicting flow at $1/8$th of the input resolution. This low-resolution flow map is then upsampled to full resolution. Upsampling methods for optical flow have different requirements than traditional image upsampling methods like bilinear interpolation. \ie, if two objects at an edge move in different directions, interpolating them in the middle will give values around the mean of the two directions. This is never correct in terms of optical flow; a pixel either follows one object, or the other. Alternatively, upsampling methods that do not use interpolation like nearest neighbor upsampling makes flow edges jagged and not aligned with object edges.

\begin{figure}[t]
    \begin{center}
        \includegraphics[width=0.6\linewidth]{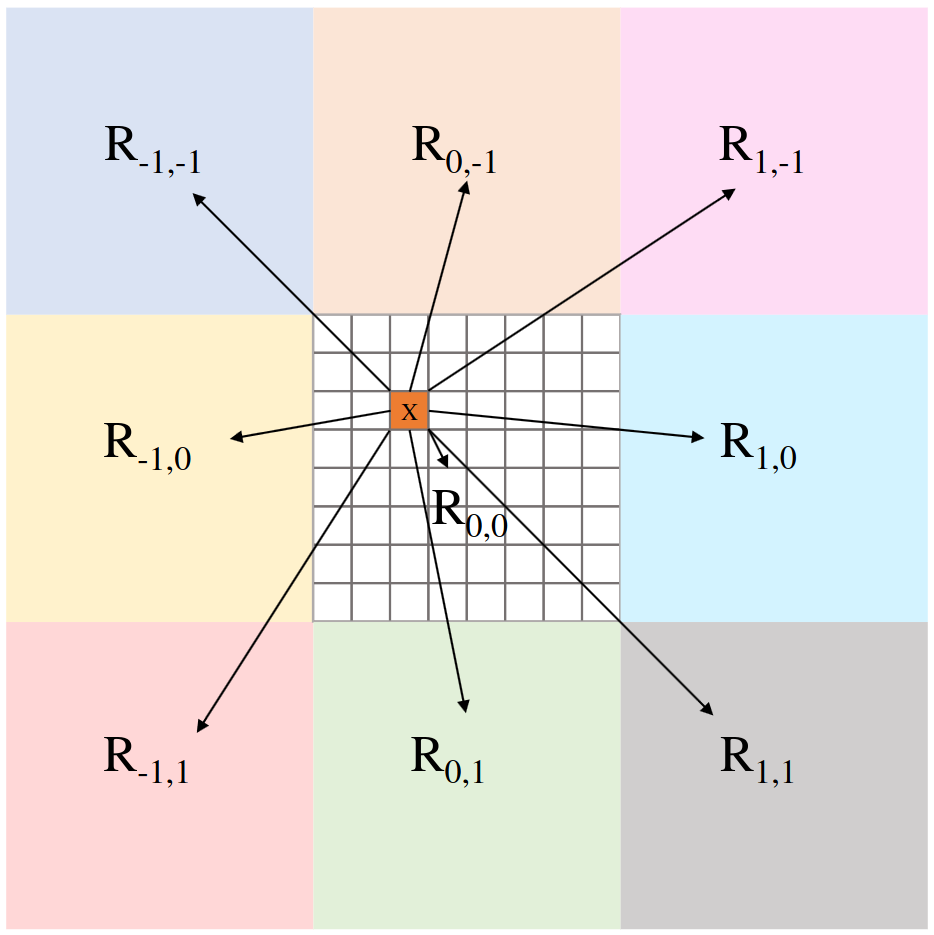}
    \end{center}
    \caption{
    Original convex optical flow upsampling as proposed by RAFT \cite{raft}. A high-resolution sub-pixel value is written as a convex combination of the low-resolution input. The convex mask weights are guaranteed to be positive and sum to $1$ by using softmax. This convex upsampling is adopted by the state of the art. We propose improvements based on the observation that the convex combination through a softmax can be rephrased as neighborhood attention~\cite{hassani2022neighborhood}, leading to increased accuracy, which can be used as a drop-in replacement for all existing models.  
    }
    \label{fig:convex_orig}
\end{figure}

To solve these issues for optical flow, RAFT~\cite{raft} proposes convex upsampling. There, the main idea is to upsample by a convex combination of the low-resolution neighbors: each low-resolution pixel is weighted, where all weights must be positive and sum to $1$, which is easily obtained by applying the softmax function, see \ref{fig:convex_orig}. Following the success of RAFT~\cite{raft}, this convex upsampling is currently adopted practically unchanged in all current SOTA flow models \citesota{}. 

In this paper, we make the observation that convex upsampling for optical flow has received relatively little attention in the literature: the design of convex upsampling has not changed much since RAFT~\cite{raft}.

Here, we rephrase the traditional convex upsampling in a more flexible model using Neighborhood Attention (NA)~\cite{hassani2022neighborhood}. Neighborhood Attention is a natural model for convex upsampling, as it is also local, and by the softmax operator inherently provides a convex combination. One of the benefits is that NA allows to decouple the number of learnable parameters from the mask size, allowing larger input sizes. 

We have the following contributions. We rephrase convex upsampling as Neighborhood Attention (NA). NA allows us to evaluate larger mask sizes, making more solutions possible. It also allows us to replace the 1-step 8x upsampling with three hierarchical steps of 2x upsampling, which helps retain spatial information. The hierarchical upsampling allows us to also include the input image features at matching resolutions, to better align flow with object edges. We also investigate decoupling the final upsampling model from the intermediate upsamplings used in the recurrent optimization. As a final investigation we explore the role of optical flow sampling in data-augmentation.

Our proposed ideas could theoretically be applied to almost every current state-of-the-art optical flow architecture. On the FlyingChairs + FlyingThings3D training setting we reduce the Sintel Clean training end-point-error of RAFT from $1.42$ to $1.26$, GMA from $1.31$ to $1.18$, and that of FlowFormer from $0.94$ to $0.90$, by solely adapting the convex upsampler. We will make all code available.

%% file: chapters/related_work.tex
\section{Related Work}
\label{related}

\paragraph{Optical Flow} is typically estimated between two consecutive frames by matching image-1 to image-2, and the seminal RAFT~\cite{raft} approach inspired much follow up work. RAFT uses a gated recurrent unit in a step-wise, recurrent, optimization process. Because of its computational complexity, the recurrent optimization is done on $1/8$th of the input image resolution, after which the low resolution output flow is upsampled to full resolution. The follow up work done by GMA~\cite{gma} adds global attention on the input features to add global, contextual information. Similarly, SeparableFlow~\cite{separableflow} improves the correlation volume construction by first ensuring global contextual features. This idea is extended by FlowFormer(++)~\cite{flowformer, flowformer_plusplus}, CRAFT~\cite{craft} and GMFlowNet~\cite{gmflownet} who add strong Transformer blocks. Essentially, RAFT is extended by improving and replacing different components other than the step-wise optimization, as this optimization approach remains the global design. So, the current optical flow state of the art~\citesotanogmflow{} is based on RAFT and inherits its main properties: low resolution iterative optimization followed by upsampling. Here, we focus on these properties and on the low resolution upsampling in particular.

MS-RAFT(+) \cite{msraft_plus, msraft} also explores different resolutions in a setting based on RAFT~\cite{raft}. Important for us is that the convex upsampling is different, as they upsample $3$ times by a factor of $2$, rather than once by a factor of $8$.
Inspired by this, we investigate the impact of this $3$-step approach when used on the other state-of-the-art networks that currently use a single upsampling operation by factor $8$.

\paragraph{Upsampling} is fundamental in computer vision tasks. Non learning-based approaches such as Nearest Neighbor, bilinear, or bicubic interpolation \cite{digital_image_proc_gonzalez, free_cv_book_szeliski}, and learning-based approaches such as Transposed convolutions~\cite{transpose_conv}, PixelShuffle as used in super resolution~\cite{sub_pix}, or convex upsampling \cite{raft}. Learned-based upsampling can be trained end-to-end \cite{transpose_conv, sub_pix, raft} or progressive as done in GANs~\cite{progressive_gans}. For optical flow, bilinear- or Nearest Neighbor interpolation can be used for decent performance~\cite{flyingchairs_flownet,flownet2,spynet} as long as the upsampling factor is small. All the current SOTA optical flow models \citesota{} require upsampling with factor $8$. This makes traditional upsampling unsuitable, and hence convex upsampling is widely adopted by these models. Here, we extend this reasoning and formulate convex upsampling as local self-attention~\cite{hassani2022neighborhood}.

\paragraph{Attention and Local Attention}
Self-attention~\cite{attention_all_you_need} used for image recogition~\cite{vision_transformers} research is  bringing back the hierarchical structure of convolutional neural networks. As such, SwinFormer~\cite{swinformer} was proposed. 
In a similar way, Neighborhood Attention (NA) \cite{hassani2022neighborhood} was proposed. While the general idea of NA is similar to SwinFormer~\cite{swinformer}, the main design difference is how the local attention maps relate to the queries. NA ensures that as long as a query is not near the image border, the query is always at the center of the local attention map. This property makes these local attention maps effectively similar to convex upsampling masks. 
An advantage of local attention maps is that their size is decoupled from the number of parameters, so we investigate the use of attention maps when taken directly as a drop-in replacement for convex upsampling masks. 

%% file: chapters/method.tex
\section{Method}
\label{method}

\begin{figure*}[t]
    \begin{center}
        \includegraphics[width=\linewidth]{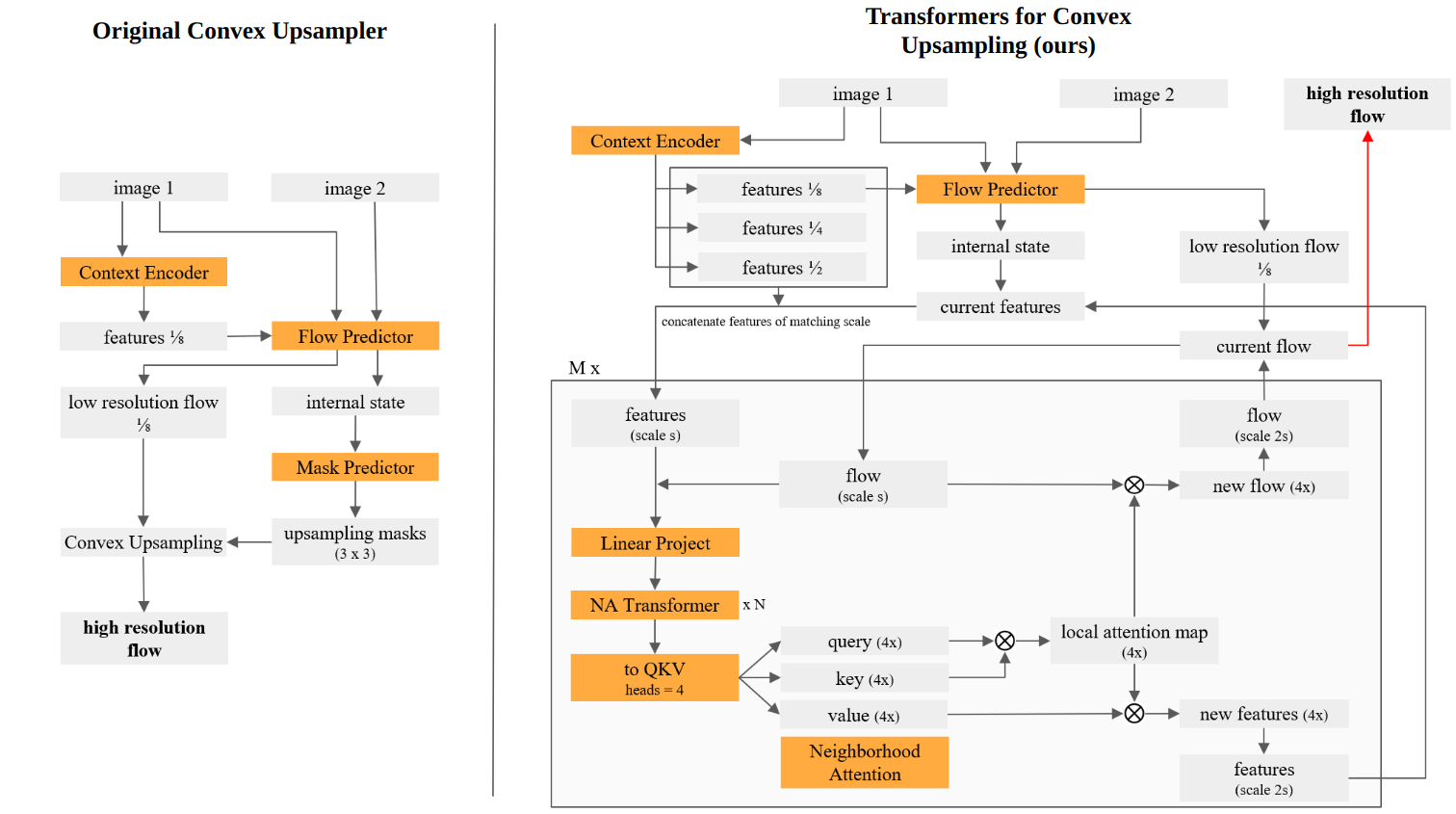}
    \end{center}
    \caption{
    \textbf{Left:} the original convex upscaling method as proposed by RAFT \cite{raft}. \textbf{Right:} our proposed multi-step Transformer convex upsampling network. The feature extractor is adopted from RAFT \cite{raft} but we extract the features at $3$ all intermediate scales. The Neighborhood Attention Transformer blocks \cite{hassani2022neighborhood} perform local neighborhood attention to enhance features. Attention is used to both upsample the low-resolution flow as well as the feature maps. The block at the bottom is executed $M$ times and upsamples by factor $2$, and hence for $M=3$ the low-resolution flow is upsampled by factor $8$.
    }
    \label{fig:design}
\end{figure*}

We show a visual comparison of the baseline convex upsampling of RAFT~\cite{raft} versus our proposed method in \ref{fig:design} and will explain its modules in the following. 

\subsection{RAFT's convex upsampler}
\label{methodRAFT}

\textbf{Context encoder}
Most SOTA models for optical flow use a context encoder that takes image-1 as input, generally based on a three-scale ResNet~\cite{resnet} architecture where each scale has a neural network $r_l$ with $l \in (0, 1, 2)$ that down-scales the resolution by factor $2$, as follows:
\vspace{-2mm}
\begin{align}
    \text{image}_{1/2} = r_0(\text{image-1})  \in \mathbb{R}^{(64 \times (H/2) \times (W/2))} \label{eq:context_12} \\
    \text{image}_{1/4} = r_1(\text{image}_{1/2})  \in \mathbb{R}^{(96 \times (H/4) \times (W/4))} \label{eq:context_14}\\
    \text{image}_{1/8} = r_2(\text{image}_{1/4})  \in \mathbb{R}^{(128 \times (H/8) \times (W/8))} \label{eq:context_18}
\end{align}
\textbf{Flow predictor}
The flow predictor takes $\text{image}_{1/8}$ and uses a linear projection to obtain the initial gated recurrent unit (GRU) internal state $h_0$, as well as the shared input to each GRU step $a$. The GRU module is defined as $g$, and the optical flow map as $\text{flow}$. We define for image input size $H \times W$ that $h = (H / 8)$ and $w = (W / 8)$.
\vspace{-2mm}
\begin{align}
    h_0 = \text{Conv}_{\text{1x1}}(\text{image}_{1/8})  \in \mathbb{R}^{(128 \times h \times w)} \label{eq:inputs_hiddeninit}\\
    a = \text{Conv}_{\text{1x1}}(\text{image}_{1/8})  \in \mathbb{R}^{(128 \times h \times w)} \label{eq:inputs_injectval}\\
    \text{flow}_0 = 0_{(2 \times h \times w)}  \in \mathbb{R}^{(2 \times h \times w)} 
\end{align}
The correlation volume $c(p)$ in RAFT \cite{raft} is defined to provide the correlation information for a given pixel at spatial position $p$. For $I$ refinement iterations the optimization approach is then defined as following, where $\text{flow}_{(I-1)}$ is the final low-resolution output flow.
\vspace{-2mm}
\begin{align}
    \text{flow}_1, h_1 = g(h_0, a, \text{flow}_0, c(\text{flow}_0)) \label{eq:gru_start} \\
    \text{flow}_2, h_2 = g(h_1, a, \text{flow}_1, c(\text{flow}_1)) \\
    \notag ... \\
    \text{flow}_{(I-1)}, h_{(I-1)} = \\ \label{eq:gru_end}
    \notag g(h_{(I-2)}, a, \text{flow}_{(I-2)}, c(\text{flow}_{(I-2)}))  \\
    \notag \text{with} \\
    \text{flow}_{(I-1)} \in \mathbb{R}^{(2 \times (H/8) \times (W/8))} \\
    h_{(I-1)} \in \mathbb{R}^{(128 \times (H/8) \times (W/8))}
\end{align}

\textbf{Mask predictor}
Now take $\text{flow}_j$ to be a low-resolution flow map with hidden state $h_j$ where $j \in (0, 1, ..., (I-1))$. For RAFT's \cite{raft} upsampler which is used to upsample the $\text{flow}_j$ from size $(h \times w)$ to $(H \times W)$, a convex combination of the low-resolution nearby pixels from $\text{flow}_j$ is used. That is, for every low-resolution pixel $P \in \text{flow}_j$, find sub-pixel values $p_i$ by first predicting a $3 \times 3$ scalar $\text{mask}_{p_i}$ for every sub-pixel $p_i$. This $\text{mask}_{p_i}$ has $9$ scalars values; $\text{mask}_{p_i} = (w_0, w_1, ..., w_{(m^2-1)})$ for mask size $m=3$. For a mask centered on low resolution pixel $P \in \text{flow}_j$, all the values within the mask including the pixel itself, form the neighbors of $P$; called $N_P \in \text{flow}_j$. Then, the value of the sub-pixel $p_i$ is calculated using the dot-product as following, for upsampling factor $f$ and mask size $m$. Furthermore, $\sigma$ refers to the use of the ReLU \cite{relu} activation function.
\vspace{-2mm}
\begin{align}
    \text{masks} = \text{Conv}_{\text{1x1}}(\sigma(\text{Conv}_\text{3x3}(h_j))) \in \mathbb{R}^{(f^2m^2)} \label{eq:fully_connected}\\
    \text{masks} \in \mathbb{R}^{(f^2 \times m \times m)} = \text{masks} \in \mathbb{R}^{(f^2m^2)} \\
    \notag i \in (0, 1, ..., (f^2 - 1)) \\
    (\text{mask}_{i}, N_{P}) \in \mathbb{R}^{(m \times m)} \\
    p_i = \langle \text{softmax}(\text{mask}_{i}), N_{P} \rangle \in \mathbb{R}^1 \label{eq:convex_up_mask_dot}
\end{align}

\subsection{Neighborhood Attention Transformers for Convex Upsampling}
For our attention-based convex upsampler, we start with the same input $h_j$, but also concatenate $\text{image}_{1/8}$ and $\text{flow}_{j}$, to build the embedding vectors $e$ for all pixels $P \in \text{flow}_{j}$ as following.
\vspace{-2mm}
\begin{align}
\text{cat} = \text{Cat}(h_{{j}}, \text{image}_{{1/8}}, \text{flow}_{{j}}) \label{eq:concat} \\
\notag \in \mathbb{R}^{(258 \times h \times w)}\\
e = \text{Conv}_{\text{1x1}}(\text{cat})  \in \mathbb{R}^{(D \times h \times w)}
\end{align}
Then we apply local Neighborhood Attention Transformers (NAT) \cite{hassani2022neighborhood} for local contextual feature enhancement, using $2$ Transformer blocks which sequentially form $T$. For these Transformer blocks we use dimensionality $D=128$ for scale $1/8$, $D=64$ for scale $1/4$, and $D=32$ for scale $1/2$. We set the head dimensionality to $32$. Note that the tokens from the embedding vectors $e$ correspond to the low-resolution pixels $P \in \text{flow}_{j}$.
\vspace{-2mm}
\begin{align}
    e_{\text{ctx}} = T(e) \in \mathbb{R}^{(D \times h \times w)}
\end{align}
Then, for upsampling factor $f$, multi-head attention is used to form $f^2$ attention maps for each $P$ $(h \times w)$ from $e_{\text{ctx}} \in \mathbb{R}^{(D \times h \times w)}$. We use head dimensionality of $\frac{1}{2}D$ such that we halve the size of the embedding dimensionality for each upsampling operation. For $f^2$ heads, this requires a query, key, and value dimensionality of $\frac{1}{2}f^2D$, which can then be reshaped to obtain $f^2$ query, key, and value maps with dimensionality $\frac{1}{2}D$.

\vspace{-2mm}
\begin{align}
    Q = \text{Conv}_{\text{1x1}}(e_{\text{ctx}}) \in \mathbb{R}^{(\frac{1}{2}f^2D \times h \times w)} \\
    K = \text{Conv}_{\text{1x1}}(e_{\text{ctx}}) \in \mathbb{R}^{(\frac{1}{2}f^2D \times h \times w)} \\
    V = \text{Conv}_{\text{1x1}}(e_{\text{ctx}}) \in \mathbb{R}^{(\frac{1}{2}f^2D \times h \times w)} \\
    (Q, K, V) \in \mathbb{R}^{(f^2 \times \frac{1}{2}D \times h \times w)}
\end{align}
Then, we construct the local attention maps using neighborhood attention (NA) \cite{hassani2022neighborhood} with window size $m$. Note that to obtain these local attention maps (LAM) they are normalized using the softmax~\cite{softmax} function, so similarly to convex upsampling masks; the values are positive and sum to $1$.
\begin{align}
    \text{LAM} = \text{NA}_{\text{maps}}(m, Q, K) \in \mathbb{R}^{(4 \times m \times m \times h \times w)}
\end{align}
We then aggregate these local attention maps with the low-resolution $\text{flow}_{j}$, as well as the value features $V$. Again, we use neighborhood attention \cite{hassani2022neighborhood} for this.
\vspace{-2mm}
\begin{align}
    \text{flow}_{\text{up}} = \text{Aggregate}(\text{LAM}, \text{flow}_{j}) \label{eq:up_flow} \\
    \text{h}_{\text{up}} = \text{Aggregate}(\text{LAM}, V) \label{eq:up_features}
\end{align}
Note that this aggregation operation on a per-pixel level is implemented as following, when $\text{mask}_{i} \in \text{LAM}_{\text{pre}}$ where $\text{LAM}_{\text{pre}}$ refers to the attention maps from LAM before softmax normalization.
\vspace{-2mm}
\begin{align}
    p_i = \langle \text{softmax}(\text{mask}_{i}), N_{P} \rangle \in \mathbb{R}^1 \label{eq:na_aggregate}
\end{align}
So, aggregation as used in local attention is effectively similar to convex upsampling, as Equation \ref{eq:convex_up_mask_dot} and Equation \ref{eq:na_aggregate} are equivalent. For both, the dot product is taken within a sliding window between the convex maps, or attention maps, and the low-resolution flow, or the values. For simplicity, we ignore the different padding implementation, as this is not necessarily relevant here.

From Equation \ref{eq:up_flow}, we obtain $\text{flow}_{\text{up}}$, which is the by factor $f$ convex upsampled flow. We also obtain the by factor $f$ upsampled features $h_{\text{up}}$ based on Equation \ref{eq:up_features}.

\paragraph{Hierarchical upsampling}
\label{hierarchical}
In the convex upsampling implementation of RAFT as described in Section \ref{methodRAFT}, upsampling is done once by factor $8$. The most important reason for this is that convex upsampling expects a feature map and a low-resolution flow map as inputs. However, it only provides a single output, namely the upsampling masks; see Equation \ref{eq:fully_connected}. Therefore, the operation cannot easily be repeated. For our TCU, this is different. As we can see from Equations \ref{eq:up_flow} and \ref{eq:up_features}; we have the same inputs as outputs, but at a different scale. We can effectively repeat this operation as often as needed, and are able to upsample $3$ times by factor $2$, rather than once by factor $8$. This has several advantages, one is a spatial inductive bias; the network only has to learn the alignment of $4$ sub-pixels at the time, rather than $64$ at once.

\paragraph{Adding the input image as extra information}
Another advantage of the hierarchical approach as discussed in the previous paragraph is that it allows for concatenating feature maps from different scales. Note that from Equations \ref{eq:context_12} till \ref{eq:context_18} we have image features at $3$ scales, yet only use the $1/8$ scale from Equation \ref{eq:context_18} as only these are used as inputs in equations \ref{eq:inputs_hiddeninit} and \ref{eq:inputs_injectval}. If we adopt our hierarchical method, we are able to concatenate each of the scales from Equations \ref{eq:context_12} ($1/2$), \ref{eq:context_14} ($1/4$) and \ref{eq:context_18} ($1/8$), to the upsampling step of our convex upsampling with corresponding scale, as could be done in Equation \ref{eq:concat}.

\paragraph{Larger mask size}

For the traditional convex upsampler, it is difficult to increase the mask size $m$ due to the use of a single vector of size $m^2f^2$ for the mask values, with mask size $m$ and upsampling factor $f$. This is because Equation \ref{eq:fully_connected} uses a fully-connected layer for this purpose, and hence increasing $m$ comes with a strong increase in the number of parameters, and is difficult to optimize. Fortunately, our Transformer based Convex Upsampler (TCU) does not have this problem. As the convex masks are simply local attention maps, and the size of local attention maps is not dependent on the number of parameters, we can freely increase the mask size $m$, as long as it fits in memory.

We use an increased mask size to explore finding new solutions. A strong limitation of convex upsampling is that a high-resolution pixel can only be predicted correctly if there exists a convex combination of the low-resolution neighbor pixels $N_P$ such that the dot-product forms the desired output value. Therefore, if the low-resolution flow map $N_{P}$ is not correct or not locally informative enough and no such convex combination exist, the desired output value can never be obtained. 
We propose to increase the size of $N_{P}$ and include more low-resolution pixels, rather than just look at the direct neighbors using a $3 \times 3$ mask. 
For our sequential upsampling steps we use mask sizes $(9, 7, 5)$, in that order from low- to high-resolution.

\paragraph{Decoupling the upsampling}
Many flow prediction architectures \citesotanogmflow{} take a recurrent step-wise refinement approach to solve optical flow, as we describe in Equation \ref{eq:gru_start} till \ref{eq:gru_end}. During training, a loss value is calculated for each intermediate flow map $\text{flow}_0, \text{flow}_1, ..., \text{flow}_{(I-1)}$, where each flow map requires to be upsampled to high resolution in order to compare it to the ground truth. Many SOTA works \citesotanogmflow{} choose to share the same convex upsampler and its weights over all steps. However, we consider the first flow predictions to be extremely noisy variants of the final flow. When the convex upsampler and its weights are shared over all steps, this is effectively equivalent to just adding strong noise to a part of the input data. In general, the loss is down-weighed for the first steps, but this could be insufficient. We want the convex upsampler of the final output refinement iteration to fully focus on its own objective, and not have its parameters shared with noisy estimates from earlier refinement iterations. To achieve this, we propose to decouple the convex upsampler of the last refinement iteration and give it its own weights.

This also brings the advantage that a different upsampling method can be used for the last refinement iteration. Our Transformed-based convex upsampling approach, TCU, uses more memory than the original convex upsampling approach. So, we exploit the idea of a decoupled upsampler for the last refinement iteration to make our TCU model feasible in practise. Namely, the original shared convex upsampler is used for the first $(I-1)$ refinement iterations, and TCU is only used for the last refinement iteration which provides the final model output. Note that at test-time, only the upsampler of the last refinement iteration is used.

%% file: chapters/experiments.tex
\section{Effect of sampling in data-augmentation} 
In addition to the architecture change, we also investigate the training scheme. 
Noticeably, almost all recent works on optical flow are based on the same original PyTorch implementation of RAFT \cite{raft}, and all use bilinear sampling for augmentation of their training data. 
The original reason of RAFT for convex upsampling was to avoid bilinear upsampling on flow maps. However, in the current setting convex upsampling is used, but with the learning objective to predict bilinearly interpolated flow. Interestingly, to the best of our knowledge, all public top submissions to the Sintel \cite{sintel} leaderboard show bilinear interpolation artifacts in the form of white and non-crisp edges, even though these artifacts are not in the non-augmented training data. 
We explore an additional training scheme to remove bilinear interpolation artifacts. Disabling this interpolation in the augmentation is likely not a good idea, as it is used to avoid overfitting. Instead, we propose an additional training scheme; (-AUG). There, a trained model is trained for an additional 40K iterations with all interpolation-based augmentations disabled.

\section{Experiments}

\textbf{Finetuning}
\label{sec:finetune}
The only difference from the original training setting is the convex upsampler from the last refinement iteration. Training all model components from random initialization in this highly similar setting would be unnecessarily costly. Instead, all the training sessions are started with pre-trained weights for the flow predictor and the convex upsampler of the first $(I-1)$ iterations. Only the weights of the convex upsampler for the last refinement iteration are randomly initialized. For all experiments, we fine-tune for $100K$ iterations with a batch size of $3$, on the dataset that the model was last trained on. A learning rate of $1e-4$ is used for the pre-trained weights, and a learning rate of $2e-4$ is used for the untrained upsampler of the last refinement iteration. We will make all code available.

\begin{table*}[t]
\centering
\resizebox{\linewidth}{!}{%
\begin{tabular}{@{}p{8cm}p{1cm}p{0.8cm}p{0.8cm}p{0.8cm}p{0.8cm}p{0.8cm}p{0.8cm}p{0.8cm}p{0.8cm}p{0.8cm}p{0.8cm}p{0.8cm}p{0.8cm}p{0.8cm}p{0.8cm}p{0.8cm}p{0.8cm}p{0.8cm}p{0.8cm}@{}}
Statistic \hspace{5cm} Bucket & 0 & 1 & 2 & 3 & 4 & 5 & 6 & 7 & 8 & 9 & 10 & 11 & 12 & 13 & 14 & 15 & 16 & 17 & 18 \\ 
\toprule
number of samples & 290,817 & 11,241 & 12,280 & 19,340 & 11,034 & 6,662 & 4,896 & 2,985 & 2,190 & 1,441 & 1,018 & 721 & 558 & 347 & 281 & 168 & 196 & 111 & 434  \\
samples (percentage) & 79.3\% & 3.1\% & 3.4\% & 5.3\% & 3.0\% & 1.8\% & 1.3\% & 0.8\% & 0.6\% & 0.4\% & 0.3\% & 0.2\% & 0.2\% & 0.1\% & 0.08\% & 0.05\% & 0.05\% & 0.03\% & 0.12\%  \\
samples (reverse cumulative percentage) & 100\% & 20.7\% & 17.6\% & 14.3\% & 9.0\% & 6.0\% & 4.2\% & 2.8\% & 2.0\% & 1.4\% & 1.0\% & 0.8\% & 0.6\% & 0.4\% & 0.3\% & 0.2\% & 0.2\% & 0.15\% & 0.12\% \\
contribution to error (percentage) & 44.0\% & 4.1\% & 5.1\% & 10.7\% & 8.6\% & 5.9\% & 5.0\% & 3.8\% & 3.1\% & 2.1\% & 1.6\% & 1.3\% & 0.9\% &  0.7\% & 0.5\% & 0.3\% & 0.5\% & 0.3\% & 1.6\% \\
contribution to error (reverse cumulative percentage) & 100\% & 56.1\% & 52\% & 46.9\% & 36.2\% & 27.6\% & 21.7\% & 16.7\% & \textbf{12.9\%} & 9.8\% & 7.7\% & 6.1\% & 4.8\% & 3.9\% & 3.2\% & 2.7\% & 2.4\% & 1.9\% & 1.6\% \\
\hline
\end{tabular}%
}
\caption{Distribution of samples over each of the buckets in Figure \ref{fig:epe_mean} for FlyingThings3D \cite{flyingthings}.}
\label{tab:sample_div}
\end{table*}

\begin{figure}[t]
    \begin{center}
        \includegraphics[width=\linewidth]{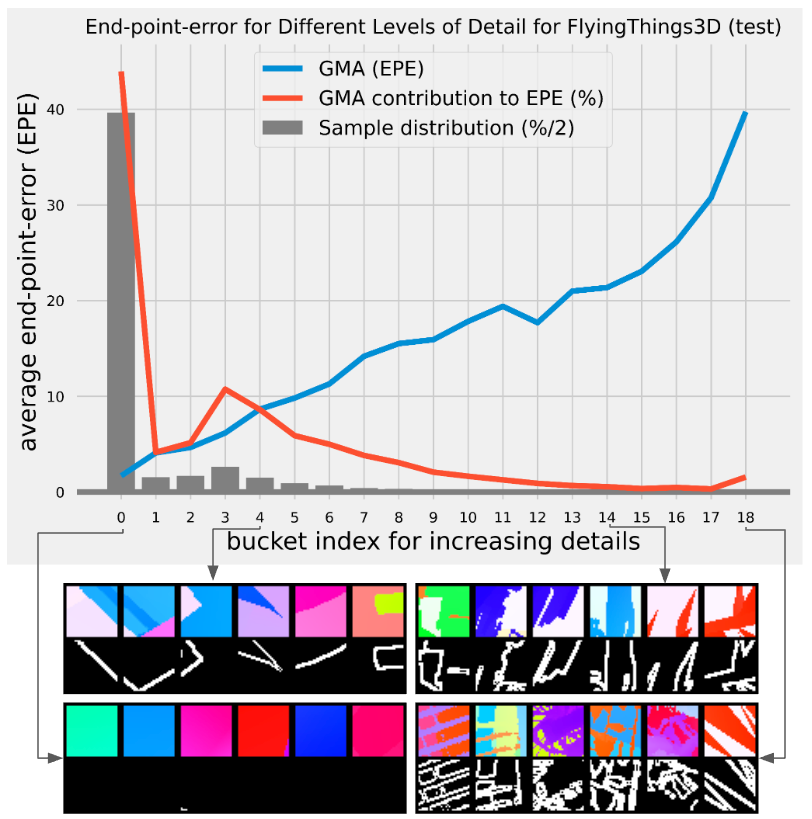}
    \end{center}
    \caption{
    The average end-point-error for increasing amount of detail on FlyingThings3D (test) \cite{flyingthings}, after being trained on C+T. The performance degrades fast for higher amount of detail. To give an impression of the level of detail per bucket, $6$ random patches from each bucket are shown for FlyingThings3D \cite{flyingthings}. The contribution of each level of detail to the overall EPE is given as well. Although it is not provided here, the graph for the Sintel \cite{sintel} dataset looks almost similar.
    }
    \label{fig:epe_mean}
\end{figure}

\subsection{Performance on High-Detail Areas}
\label{sec:binary_detail}

An important reason for our upsampler is the performance on high-detail areas. To investigate this, we look at the performance on non-overlapping $32 \times 32$ patches of the test data. We take the ground truth optical flow and assume the number of edge pixels in the ground truth flow map to strongly correlate with the amount of details, and hence with the difficulty for the convex upsampler. We extract spatial gradients using the Kornia library \cite{kornia}, and consider the $L_2$ norm on the $4$-dimensional vector that comes from extracting the $\partial x$ and $\partial y$ gradients for each $XY$ channel as an edge detector. To obtain a binary edge map, a binary threshold of $8$ is used. To get the level of detail for a patch, the average value is taken of the binary edge map of that patch. Next, we plot the average end-point-error (EPE) for each level of detail, for which a bin width of $0.02$ is used. Samples with level of detail that is larger than the specified domain are placed in the last bucket. 

The results in Figure \ref{fig:epe_mean} show a strong degrade in accuracy for increasing amounts of detail, which confirms that our hypothetical problem exists. We provide the statistics on the amount of samples per bucket from Figure \ref{fig:epe_mean} in Table \ref{tab:sample_div}. From this table, note for example that the buckets $(b >= 8)$ only contain 2\% of the patches, yet contribute for 13\% to the end-point-error. This strongly highlights the importance of our method, as even though the amount of high-detail patches is low, its impact on the end-point-error can be significant.

\subsection{Transformers for Convex Upsampler}
\label{sec:model_names}
Next, we investigate the impact of the proposed individual components. \textbf{+DC} refers to decoupling the last refinement's convex upsampler, and giving it its own weights. \textbf{+FT} refers to concatenating the features from the context branch to the input of the convex upsampler. When TCU is used features are added at all scales, otherwise only the low resolution features are appended to the input. \textbf{+TCU(a/b/c)} refers to the use of our Transformers for Convex Upsampling (TCU) with mask size $a$ for the first upsampling step, $b$ for the second step, and $c$ for the last step. Lastly, \textbf{-AUG} refers to the additional fine-tuning steps with disabled interpolation-based augmentations.

\begin{table}[t]
\resizebox{\linewidth}{!}{
\begin{tabular}{@{}p{3cm}p{1cm}p{1cm}p{1cm}p{1cm}p{1cm}p{1cm}p{2cm}@{}}
\multirow{2}{*}{ Method } & \multicolumn{2}{c}{ FlyingThings3D } & \multicolumn{2}{c}{ Sintel (train) } & \multicolumn{2}{c}{ KITTI-15 (train) } & \multirow{2}{2cm}{ Number of parameters } \\
& Train & Test & Clean & Final & F1-epe & F1-all &  \\ \hline
\multicolumn{1}{l}{GMA (recomputed)} & 10.34 & 3.07 & 1.31 & \textbf{2.75} & 4.48 & 16.86 & 443K \\
\multicolumn{1}{l}{+DC} & 9.38 & 2.84 & 1.23 & 2.78 & 4.43 & 16.89 & 443K  \\
\multicolumn{1}{l}{+DC+FT} & 9.53 & 2.86 & 1.24 & 2.79 & 4.55 & 16.92 & 743K \\
\multicolumn{1}{l}{+DC+TCU(3/3/3)+FT} & 9.51 & 2.73 & 1.22 & 2.83 & 4.52 & 16.80 & 695K  \\
\multicolumn{1}{l}{+DC+TCU(9/7/5)+FT} & 9.24 & 2.75 & 1.21 & 2.80 & \textbf{4.36} & \textbf{16.26} & 700K  \\
\multicolumn{1}{l}{+DC+TCU(9/7/5)+FT-aug} & \textbf{8.97} & \textbf{2.61} & \textbf{1.18} & 3.01 & 4.50 & 16.64 & 700K  \\ \hline
\end{tabular}
}
\caption{Training and generalization performance for our different convex upsampling approaches. Model names are explained in Section \ref{sec:model_names}.}
\label{tab:small_comp}
\end{table}

\begin{figure*}[t]
    \begin{center}
        \includegraphics[width=0.32\linewidth]{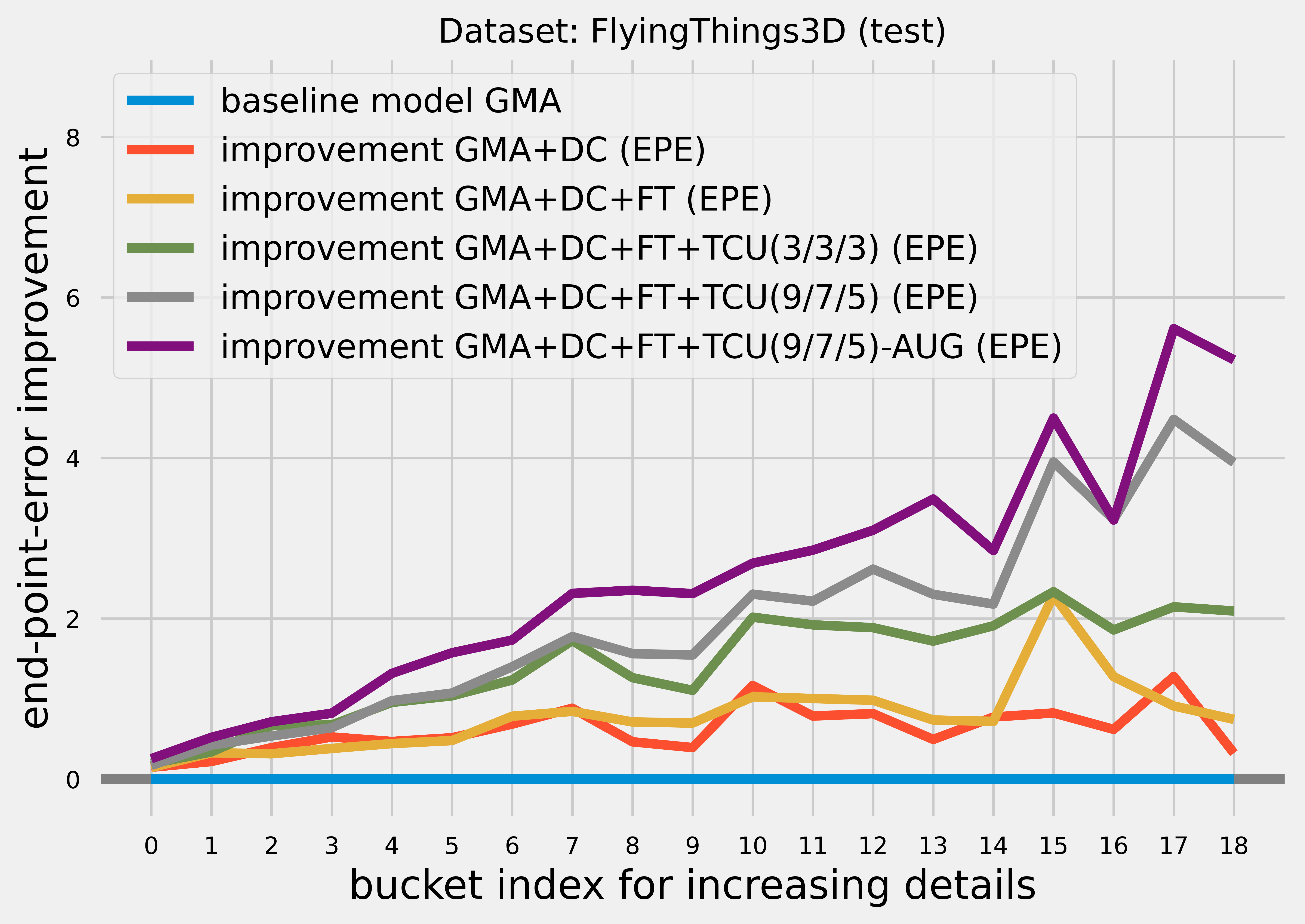}
        \includegraphics[width=0.32\linewidth]{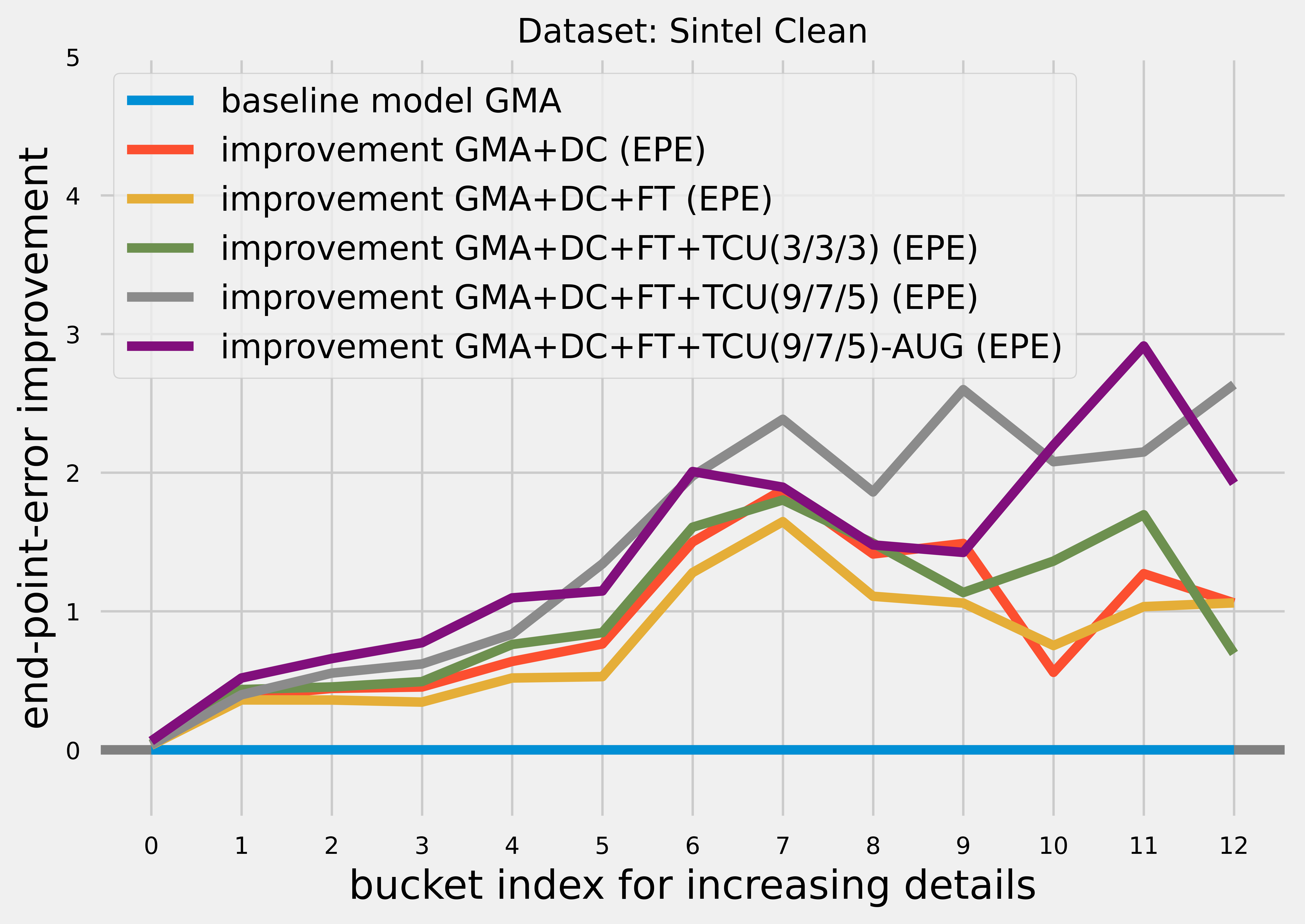}
        \includegraphics[width=0.33\linewidth]{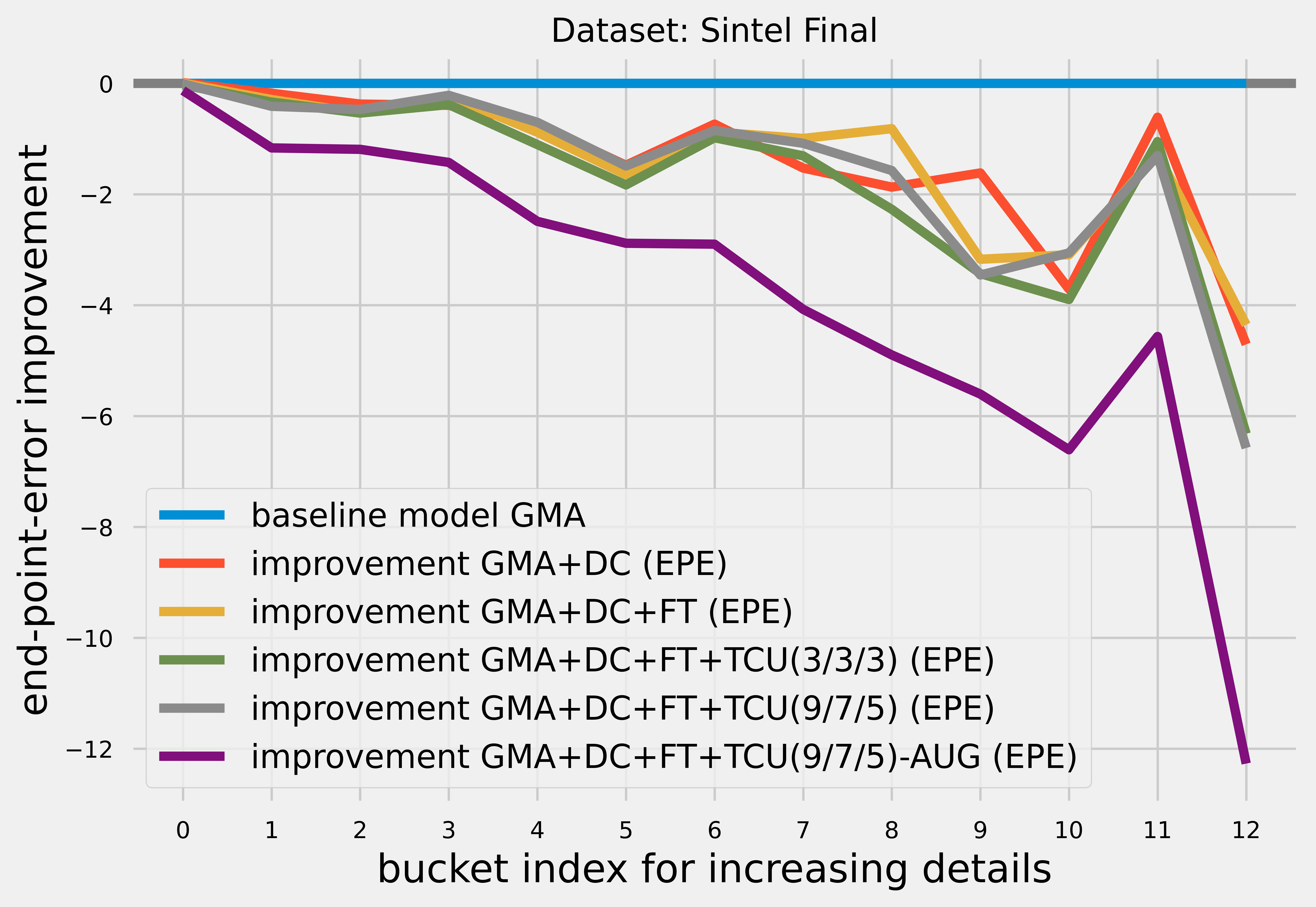}
    \end{center}
    \caption{
    Improvement in end-point-error for increasing amount of detail on FlyingThings3D (test) \cite{flyingthings}, Sintel Clean (train) and Sintel Final (train) \cite{sintel} , for a series of our proposed models, after being trained on FlyingChairs (train) \cite{flyingchairs_flownet} and FlyingThings3D (test) \cite{flyingthings}.
    }
        \label{fig:pm_ft3d}
        \label{fig:pm_sintel_final}
        \label{fig:pm_sintel_clean}
\end{figure*}

\paragraph{FlyingThings3D}
We first investigate the performance for several of our proposed models on the FlyingThings3D \cite{flyingthings} test data, after being trained on C+T. The results hereof are shown in Figure \ref{fig:pm_ft3d}. From this figure we observe that all our proposed changes result in improvements over the previous model that did not have the change. This is as expected, as we do not expect strong relations between each of our proposed changes, as each of our proposed changes aims at providing improvements in a different way. While these results are good, it is also important to consider situations where this might not be the case. This result is calculated for the test data of the dataset that the model was trained on, so there is no measure of cross-dataset generalization here. For cross-dataset generalization performance, we consider the performance on Sintel Clean \cite{sintel}, which is done next.

\paragraph{Sintel Clean}

We create the same graph as before, but now for the Sintel Clean \cite{sintel} dataset. The results hereof are shown in Figure \ref{fig:pm_sintel_clean}. In general, similar patterns as for FlyingThings3D \cite{flyingthings} are observed. However, the gap between with and without (-AUG) is no longer as clear as before. Possibly, interpolation artifacts on edges is a good thing for cross-dataset generalization. Reason for this could be that interpolation on edges results in the `safe choice` for a model as it provides values around the mean, which then on average provides similar performance as sharp edges that are sometimes wrong, when evaluated on the end-point-error.

An interesting result from this is that increasing the mask size for TCU from (3/3/3) to (9/7/5) again provides clear improvements. This, together with the same result for FlyingThings3D \cite{flyingthings}, provides empirical evidence for our hypothesis that a larger mask size can makes more convex solutions possible, which in turn improves performance.

\paragraph{Sintel Final}
For Sintel Final \cite{sintel} there exist some important differences. Mainly, Sintel Final contains many strong blurring-based effects in an attempt to mimic real-world camera effects such as motion blur. This introduces a very important aspect; aligning flow with image edges is not a good thing. For Sintel Final, we would instead like to have a model that learns where the actual object edges are based on an image that contains motion blur. To inspect the performance, the same graph as before is generated, but now for Sintel Final \cite{sintel}. The results hereof can be found in Figure \ref{fig:pm_sintel_final}. As expected, every step we take towards better aligning flow with objects edges, degrades the models performance for this dataset. Interestingly, removing the bilinear interpolation artifacts from the model output (-AUG) causes a steep decrease in model performance. Clearly, bilinear interpolation artifacts provide an advantage here. This again confirms our earlier idea that possibly bilinear interpolation artifacts on edges are a `safe choice` when evaluated with the end-point-error, as the values are around the mean. If this is indeed the case, it makes sense that these artifacts help for strong cross-dataset generalization where edges do not align with flow.

Overall, our GMA+DC+FT+TCU(9/7/5) model sets a strong new baseline on all datasets, except Sintel Final. This is shown in Table \ref{tab:small_comp}. While cross-dataset generalization is an important aspect of optical flow models, we do not believe it to be realistic to build a model that generalizes to such impactful motion-blur artifacts that are not at all in the training data. It is important to note that generalization to KITTI-15 \cite{kitti} is good, even though it consists of actual real-world images, taken by a camera.

\input{images/large_table.tex}

\paragraph{General Comparison} results are reported in Table \ref{tab:large_scale} for various settings (\eg C+T, C+T+S+K+H). While the C+T setting can be seen as a good measure of cross-dataset generalization, training with the training data of specific datasets is also considered interesting. Therefore, we integrate our approach on GMA \cite{gma} for the C+T+S+K+H setting. As we observe here, our method can provide an improvement on Sintel Final as well when the training data also contains these blurring artifacts, as is the case for this setting.

%% file: images/large_table.tex
\newcommand{\ul}{\underline}
\newcommand{\tld}{$^\dagger$}

\begin{table*}[t]
\centering
\resizebox{0.8\linewidth}{!}{%
\small
\begin{tabular}{@{}p{2cm}p{1cm}p{1cm}p{1cm}p{1cm}p{1cm}p{1cm}p{1cm}p{2cm}@{}}
\multirow{2}{*}{ Training Data } & \multirow{2}{*}{ Method } & \multicolumn{2}{c}{ Sintel (train) } & \multicolumn{2}{c}{ KITTI-15 (train) } & \multicolumn{2}{c}{Sintel (test)} & \multicolumn{1}{c}{ KITTI-15 (test) } \\
& & Clean & Final & F1-epe & F1-all & Clean & Final & F1-all \\ \hline

\multirow{25}{*}{C+T} & \multicolumn{1}{l}{HD3 \cite{hd3} } & 3.84 & 8.77 & 13.17 & 24.0 & - & - & - \\
& \multicolumn{1}{l}{PWC-Net \cite{pwcnet} } & 2.55 & 3.93 & 10.35 & 33.7 & - & - & - \\
& \multicolumn{1}{l}{LiteFlowNet2 \cite{liteflownet2} } & 2.24 & 3.78 & 8.97 & 25.9 & - & - & - \\ 
& \multicolumn{1}{l}{VCN \cite{vcn} } & 2.21 & 3.68 & 8.36 & 25.1 & - & - & - \\
& \multicolumn{1}{l}{MaskFlowNet \cite{maskflownet} } & 2.25 & 3.61 & - & 23.1 & - & - & - \\
& \multicolumn{1}{l}{FlowNet2 \cite{flownet2} } & 2.02 & 3.54 & 10.08 & 30.0 & - & - & - \\
& \multicolumn{1}{l}{DICL-Flow \cite{dicl_flow} } & 1.94 & 3.77 & 8.70 & 23.6 & - & - & - \\
\\
& \multicolumn{1}{l}{RAFT \cite{raft} } &                   1.43 &      2.71 &      5.04 &      17.4 \\
& \multicolumn{1}{l}{\hspace{0.5cm}RAFT (recomputed) } &    1.42 &      \ul{2.69} & 5.01 &      17.5 & - & - & - \\
& \multicolumn{1}{l}{\hspace{0.5cm}RAFT+ALL (ours) } &      \ul{1.26} & 2.74 &      4.92 &      \ul{17.4} & - & - & - \\
& \multicolumn{1}{l}{\hspace{0.5cm}RAFT+ALL-aug (ours) } &  1.28 &      2.93 &      \ul{4.90} & 17.5 & - & - & - \\
\\
& \multicolumn{1}{l}{GMA \cite{gma} } &                     1.30 &      2.74 &      4.69 &      17.1 & - & - & - \\
& \multicolumn{1}{l}{\hspace{0.5cm}GMA (recomputed) } &     1.31 &      \ul{2.75} & 4.48 &      16.9 & - & - & - \\ 
& \multicolumn{1}{l}{\hspace{0.5cm}GMA+ALL (ours) } &       1.21 &      2.77 &      \ul{4.47} & 17.0 & - & - & - \\ 
& \multicolumn{1}{l}{\hspace{0.5cm}GMA+ALL-aug (ours) } &   \ul{1.18} & 3.01 &      4.50 &      \ul{16.6} & - & - & - \\
\\
& \multicolumn{1}{l}{SeparableFlow \cite{separableflow} } & 1.30 & 2.59 & 4.60 & 15.9 & - & - & - \\
& \multicolumn{1}{l}{GMFlowNet \cite{gmflownet} } & 1.14 & 2.71 & \textbf{4.24} & 15.4 & - & - & - \\
\\
& \multicolumn{1}{l}{FlowFormer \cite{flowformer} } &           1.01 &      2.40 &      4.09\tld &  14.7\tld & - & - & - \\
& \multicolumn{1}{l}{\hspace{0.5cm}FlowFormer (recomputed) } &  0.94\tld &  \ul{\textbf{2.33}}\tld &  \ul{\textbf{4.24}}\tld &  \ul{\textbf{14.9}}\tld & - & - & - \\ 
& \multicolumn{1}{l}{\hspace{0.5cm}FlowFormer+ALL (ours) } &    \ul{\textbf{0.90}}\tld &  2.34\tld &  4.57\tld &  15.4\tld & - & - & - \\ 
& \multicolumn{1}{l}{\hspace{0.5cm}FlowFormer+ALL-aug (ours) }& 0.91\tld &     2.37\tld &     4.52\tld &  15.1\tld & - & - & - \\ 

\hline

\multirow{9}{*}{C+T+S+K+H} & \multicolumn{1}{l}{LiteFlowNet2 \cite{liteflownet2} } & (1.30) & (1.62) & (1.47) & (4.8) & 3.48 & 4.69 & 7.74 \\
& \multicolumn{1}{l}{PWC-Net+ \cite{pwc_net_plus}} & (1.71) & (2.34) & (1.50) & (5.3) & 3.45 & 4.60 & 7.72 \\
& \multicolumn{1}{l}{VCN \cite{vcn}} & (1.66) & (2.24) & (1.16) & (4.1) & 2.81 & 4.40 & 6.30 \\
& \multicolumn{1}{l}{RAFT \cite{raft} } & (0.76) & (1.22) & (0.63) & (1.5) & 1.61* & 2.86* & 5.10 \\ %

\\
& \multicolumn{1}{l}{GMA \cite{gma} } & - & - & - & - & \textbf{1.39}* & 2.47* & 5.15 \\ 
& \multicolumn{1}{l}{\hspace{0.5cm}GMA (recomputed) } & (0.63) & (1.05) & \textbf{(0.58)} & \textbf{(1.3)} & - & - & - \\ 
& \multicolumn{1}{l}{\hspace{0.5cm}GMA+ALL (ours) } & (0.58) & (0.97) & (0.62) & (1.4) & 1.45* & \textbf{2.44}* & - \\ 
& \multicolumn{1}{l}{\hspace{0.5cm}GMA+ALL-aug (ours) } & \textbf{(0.55)} & \textbf{(0.90)} & \textbf{(0.58)} & \textbf{(1.3)} & 1.44* & 2.47* & \textbf{5.03} \\ %
\hline
\end{tabular}%
}
\caption{ General comparison of our proposed models against other works. Here, ALL refers to our `+DC+TCU(9/7/5)+FT` setting. Furthermore, C+T refers to training on FlyingChairs \cite{flyingchairs_flownet} and then on FlyingThings3D \cite{flyingthings}. Next, C+T+S+K+H refers to training on a mix of data from FlyingChairs \cite{flyingchairs_flownet}, FlyingThings \cite{flyingthings}, Sintel \cite{sintel}, KITTI-15 \cite{kitti}, and HD1K \cite{hd1k}. The values in parentheses `()` are calculated on training data that the model was already trained on. *The warm start strategy is used as described by RAFT \cite{raft}. $^\dagger$The tile technique is used for evaluation as described by FlowFormer \cite{flowformer}. Note that to make training feasible for FlowFormer we do not fine-tune the transformer models, and only use features from scales $1/4$ and $1/8$ as scale $1/2$ cannot be obtained.}
\label{tab:large_scale}
\end{table*}

%% file: chapters/discussion.tex
\section{Discussion}

We observe that our -AUG training scheme decreases the presence of these artifacts in the model output, as can be seen in our submission to the Sintel public scoreboard. Possibly, completely removing these artifacts would require longer training without augmentations. However, this will likely cause an overfit to the training data, so we leave a better solution to this for future work. 

Overall, we find good results by reconsidering the design of the convex upsampler and the bilinear interpolation on the flow during training. In general, all our methods; +TCU, +DC, +FT, and -AUG achieve a better fit on the the training data. This was our initial goal, and therefore we can confirm that our changes have the desired effect. However, generalization is an important aspect of optical flow models. As such, we ask for careful consideration for adopting our methods when a large generalization gap exists, as in such case our method may not result in improvements. When there is no large generalization gap present as is the case for Sintel Clean, we show in the (C+T) setting that all our proposed changes can provide improvements on a wide variety of models such as RAFT, GMA, and FlowFormer. We expect similar improvements for other models that currently use the original convex upsampling by factor $8$.

Lastly, it is important to consider the accuracy of the edges in the training data. For example, KITTI-15~\cite{kitti} has its flow maps constructed from sensory data, so its exact accuracy on minor details can possibly be off. Sintel Clean is considered a strong benchmark as the flow map is constructed with 100\% certainty, and the images form a good representation of the high-detail edges. Possibly, Spring \cite{spring} would be a good evaluation metric, but unfortunately at the time of this work, this dataset has not yet been released.